\documentclass{article}

\usepackage[main, preprint]{neurips_2026}

\usepackage[utf8]{inputenc}
\usepackage[T1]{fontenc}
\usepackage{hyperref}
\usepackage{url}
\usepackage{booktabs}
\usepackage{amsfonts}
\usepackage{amsmath}
\usepackage{amssymb}
\usepackage{amsthm}
\usepackage{nicefrac}
\usepackage{microtype}
\usepackage{xcolor}
\usepackage{graphicx}
\usepackage{float}
\usepackage{algorithm}
\usepackage{algpseudocode}

\title{Attribution-Based Neuron Utility for Plasticity Restoration in Deep Networks}

\author{                                                      
    Patrick Elisii\thanks{Corresponding author. Email: \texttt{patrick\_elisii@vanguard.com}} \qquad Lucas Beauchemin \qquad Dawer Jamshed \\
    \\
    The Vanguard Group, Inc. \\                  
}  

\begin{document}

\maketitle

\begin{abstract}
Continual learning research attempts to conserve two fundamental capabilities: new knowledge acquisition and the preservation of previously acquired knowledge. While knowledge in this case can be measured through performance over an implicit or explicit task space, model plasticity generally concerns adaptability as data distributions evolve. Though much of the literature has focused on catastrophic forgetting, deep networks can also suffer from loss of plasticity, becoming progressively harder to update under continued training. Recent research has identified multiple mechanisms underlying this phenomenon, including neuron saturation, parameter norm growth, and loss of useful curvature directions. Adaptive reset-based interventions, which selectively reinitialize low-utility network parameters, have emerged as practical solutions to restore trainability. Existing utility measures used to guide resets, such as activation magnitude, contribution utility, or gradient-based activity, rely on proxy signals that can become misaligned with the intervention they are meant to guide. In this paper, we introduce gradient times difference from reference (GXD), a theoretically motivated utility measure based on reference-based gradient attribution that estimates the first-order functional cost of replacing a unit. Our results show that utility measures aligned with the functional cost of the reset can make interventions more reliable in settings where existing reset criteria degrade. GXD reframes adaptive resetting as an intervention cost estimation problem, providing a practical path toward more robust continual learning systems.
\end{abstract}

\section{Introduction}

Continual learning is an increasingly important area of machine learning research, with applications spanning computer vision, time series forecasting, and natural language processing \citep{wang2024continual}. A major challenge of continual learning is the loss of plasticity in deep neural networks, where models trained incrementally on non-stationary data become progressively less able to adapt to new information \citep{dohare2024loss}. This failure mode is distinct from, though often intertwined with, catastrophic forgetting. Rather than only losing previously acquired knowledge, the model also loses the capacity to acquire new knowledge efficiently. Prior work has linked this degradation to dormant or saturated units \citep{sokar2023dormant}, pre-activation and target-distribution shift \citep{lyle2024disentangling}, parameter norm growth and reduced effective rank \citep{dohare2024loss,lyle2024disentangling}, and loss of useful curvature directions \citep{lewandowski2023curvature}. Because plasticity loss has been shown to arise from multiple interacting mechanisms, adaptive interventions that restore capacity during training have become an important practical direction.

Reset-based methods are one such intervention. These methods periodically reinitialize parts of the network judged to have low utility, injecting fresh capacity while attempting to preserve the current function. ReDo resets neurons with low normalized activity \citep{sokar2023dormant}, Continual Backpropagation (CBP) continuously replaces a small fraction of low-utility hidden units \citep{dohare2024loss}, Selective Weight Reinitialization extends this idea to individual weights \citep{hernandez2025selective}, and ReGraMa uses gradient information to identify neurons for recycling in reinforcement learning \citep{liu2025regrama}. These methods differ not only in how they measure utility, but also in how utility is used to allocate resets.

We focus on the utility problem in rank-based reset methods such as CBP. In these methods, the reset rate fixes how much plasticity is injected, while the utility score determines where that intervention is applied. This separates two questions that are often conflated: what is the frequency and magnitude of resets, and which units should be reset? From this perspective, unit utility is a ranking signal for allocating a fixed reset budget under a plasticity-stability tradeoff. The reset should provide plasticity benefit while minimizing disruption to the current function, thus a central role of utility is to estimate the functional consequences of replacing a candidate unit.

Existing reset utilities capture different parts of this tradeoff, but none directly estimates the downstream cost of the reset intervention. Activation-based dormancy scores measure local expression \citep{sokar2023dormant}, while CBP's contribution utility adds outgoing weights and a running reference activation \citep{dohare2024loss}. These local proxies can become unreliable when downstream influence is decoupled from local magnitude, as can occur with non-ReLU activations, normalization layers, skip connections, and multi-branch computation \citep{liu2025regrama}. Loss-gradient utilities add downstream awareness, but they rank units by their effect on the current loss rather than by the functional perturbation caused by replacing them. For rank-based resets, where the reset budget is fixed, utility should help allocate replacement toward units whose reset preserves trainability without unnecessarily disrupting the current function. We focus on the directly estimable part of this problem: the functional cost of the actual intervention, moving a unit from its current activation to the reset reference used by the reset mechanism.

We propose gradient times difference from reference (GXD), a utility score that estimates this intervention cost. GXD weights a unit's displacement from the reset reference by the sensitivity of a task-relevant output to that displacement. In the CBP setting, the reference is the running activation value used by reset compensation \citep{dohare2024loss}, so GXD estimates the first-order downstream effect of moving a unit from its current activation to the effective value it will take after reset. This connects reset utility to reference-based attribution methods such as DeepLIFT \citep{shrikumar2017learning} and Integrated Gradients \citep{sundararajan2017axiomatic}, but uses attribution not for post hoc interpretation, but as an online ranking signal for low-cost plasticity injection.

We make three contributions. First, we formulate utility estimation for rank-based reset methods as a plasticity-stability tradeoff and identify downstream reset cost as the key quantity needed to allocate a fixed reset budget. Second, we introduce GXD as a simple reference-relative attribution utility that aligns the score with the reset intervention used by CBP. Third, we evaluate GXD in settings where local utility measures become unreliable. We show that GXD better predicts realized reset shock, improves Continual Backpropagation under smooth and non-ReLU activations, mitigates plasticity loss in networks with layer normalization, and improves feature stability in residual architectures. These results support the view that reset utility should estimate the downstream functional cost of a reset, rather than only local expression, local contribution, or learning signal.

\section{Related work}

Loss of plasticity has been linked to several interacting mechanisms, including dormant or saturated units, pre-activation and target-distribution shift, parameter norm growth, reduced effective rank, and loss of useful curvature directions \citep{sokar2023dormant, lyle2024disentangling, dohare2024loss, lewandowski2023curvature}. Existing mitigations either modify global training dynamics, through normalization, weight decay, regenerative or spectral regularization, and shrink-and-perturb style noise injection \citep{ash2020warm, kumar2023maintaining, lewandowski2025spectral}, or intervene directly on network components through resets and plasticity injection \citep{nikishin2022primacy, nikishin2023plasticity, dohare2021continual, sokar2023dormant, hernandez2025selective, liu2025regrama}. Our work focuses on this second family, and more specifically on the utility signal used to choose which units a rank-based reset method should replace. Prior reset utilities rely mainly on activation statistics, activation-weight contribution heuristics, or loss-gradient activity \citep{dohare2024loss, sokar2023dormant, liu2025regrama}; in contrast, we draw on feature-attribution methods that estimate how internal components affect model outputs, including DeepLIFT, Integrated Gradients, gradient-based attribution, conductance, and efficient internal-neuron importance scores \citep{shrikumar2017learning, sundararajan2017axiomatic, ancona2018towards, dhamdhere2019important, shrikumar2018computationally}. While attribution-based importance scores have been used mainly for interpretation and pruning \citep{yeom2021pruning, yvinec2022singe}, we use attribution as an online utility estimator for low-impact reset selection.

\section{Preliminaries}

\subsection{Continual Backpropagation as rank-based reset}

We consider continual learning settings in which a network is trained on a long sequence of changing input distributions or tasks. In such settings, standard backpropagation can cause networks to progressively lose plasticity, meaning that the network becomes less able to adapt to new data as training proceeds. Continual Backpropagation (CBP) addresses this problem by augmenting ordinary gradient-based learning with a continual generate-and-test mechanism that reinitializes a small fraction of mature low-utility hidden units during training \citep{dohare2021continual,dohare2024loss}.

CBP is a rank-based generate-and-test reset method: after each train step, it updates a tracked utility for each hidden unit and replaces mature units with the lowest utilities according to a set replacement rate \(\rho\). A reset samples new incoming weights, and resets the unit's optimizer state and age. After a reset, outgoing weights are set to zero and the bias of each downstream consumer is adjusted by the removed unit's average contribution, \(w_{i,k,t}^{(l)}\hat f_{l,i,t}\), to reduce the immediate functional effect of removal.

The contribution utility used by \citet{dohare2024loss} tracks local expression weighted by outgoing connection magnitude, \(u_i^{\mathrm{Cont}}\approx\mathbb{E}[|h_i|\sum_k |w_{i,k}^{\mathrm{out}}|]\). Another proposed utility, mean-corrected adaptable contribution \citep{dohare2024maintaining}, replaces \(|h_i|\) with displacement from a running activation reference, \(|h_i-r_i|\), and includes an adaptation factor inversely proportional to incoming weight magnitude. Full CBP algorithm and pseudocode are given in Appendix~\ref{app:cbp_details}.

These utilities make CBP an ideal testbed for studying reset selection because the intervention is fixed, but different utility scores induce different rankings of which units to replace. We therefore keep the CBP reset mechanism fixed and modify only the utility estimator used to rank eligible neurons.

\subsection{Reset utility as a cost--benefit tradeoff}

Rank-based reset methods choose which mature units should receive a fixed
reset intervention. Resetting a neuron can be beneficial because it can restore
future trainability by replacing a unit whose current state limits adaptation.
However, the same reset can also be harmful because it removes learned features that
 currently support the network's represented function. Thus, reset selection may be considered as a cost-benefit problem which balances reset cost and future trainability.

Let \(R_S(\theta_t)\) denote the parameters obtained by applying the reset
operator to a set \(S\) of mature units at time \(t\). Let
\(\mathcal{D}_t\) denote the current or recent online data distribution and let
\(z_\theta(x)\in\mathbb{R}^C\) denote the logits. We define the immediate
functional cost of resetting \(S\) as
\begin{equation}
  C_t(S)
  =
  \mathbb{E}_{x \sim \mathcal{D}_t}
  \left[
    d\left(z_{\theta_t}(x), z_{R_S(\theta_t)}(x)\right)
  \right],
  \label{eq:ideal_reset_cost}
\end{equation}
where \(d\) is a task-relevant output distance, such as logit distance or KL divergence between predictive
distributions. Let \(B_t(S)\) denote the expected future plasticity benefit of
resetting \(S\): the improvement in subsequent adaptation obtained by replacing
those units and continuing training on future data. An ideal fixed-budget reset
selector would solve
\begin{equation}
  S_t^\star
  =
  \arg\max_{\substack{S \subseteq \mathcal{M}_t \\ |S|=k}}
  \left[
    B_t(S) - \lambda C_t(S)
  \right],
  \label{eq:ideal_reset_value}
\end{equation}
where \(\mathcal{M}_t\) is the set of mature reset-eligible units, \(k\) is the
reset budget, and \(\lambda\) controls the plasticity--stability tradeoff.

The two terms in Equation~\ref{eq:ideal_reset_value} differ in how directly they
can be estimated. The future benefit \(B_t(S)\) is a delayed counterfactual
quantity whose value depends on the subsequent use of the reset units after training continues.
Existing plasticity-oriented utilities, such as loss-gradient magnitude or
incoming-weight norm, therefore act as proxies for this benefit term by trying
to identify units whose current state may limit future adaptation. These signals
incentivize replacing units that receive low learning signal or are difficult to
adapt. By contrast, the cost \(C_t(S)\) has a direct interventional target at
selection time, since one can apply the same reset operation to a candidate unit
and measure how much the current function changes. Contribution-style
metrics, including CBP's original contribution and mean-corrected
contribution terms, can therefore be understood as proxies for this cost term.
They locally estimate how much the current computation depends on a unit and discourage
replacing units whose removal would perturb downstream activity.

Because unit-level plasticity benefit is delayed, online
reset methods can only incentivize it through proxies. Functional reset cost, however, has a direct interventional target. Since fixed-budget resets of poorly
chosen units can hinder training or cause performance collapse
\citep{sokar2023dormant}, we primarily focus on estimating this cost term. Section~4 then
shows that our proposed attribution score still retains a conservative
connection to trainability through the downstream Jacobian.

\subsection{Why existing utilities can mis-rank reset candidates}

Activation-based utilities measure local expression, which can be a useful dormancy signal in simple feed-forward ReLU networks. However, activation magnitude is already known to be a weak replacement-suitability proxy outside this regime, and prior work has improved on it by incorporating contribution \citep{dohare2024loss} or gradient information \citep{liu2025regrama, hernandez2025selective}.

CBP's mean-corrected contribution utility is a stronger stability proxy because it is tied to the reset compensation. Let \(r_i=\hat f_i\) be the running reference activation used by CBP. For a downstream preactivation
\[
q_k(x)=b_k+w_{i,k}h_i(x)+\sum_{j\neq i}w_{j,k}h_j(x),
\]
CBP removes unit \(i\) by setting its outgoing weights to zero and transferring \(w_{i,k}r_i\) into the downstream bias. The immediate preactivation change is therefore
\[
q_k(x)-q'_k(x)=w_{i,k}(h_i(x)-r_i).
\]
Thus, CBP's mean-corrected contribution estimates the local next-layer perturbation induced by the compensated reset. This explains why it improves over raw activation magnitude: units whose outputs are nearly constant around their recent reference can be assigned low utility even if their absolute activations are large.

The limitation is that this estimate remains local. It assumes that the measured activation is the component being reset, that the next consumer is an ordinary preactivation, and that one-hop perturbation magnitude is a faithful proxy for output impact. These assumptions can fail when branch fusion, normalization, downstream nonlinearities, or task-relevant output directions amplify, suppress, or redistribute the perturbation. Two units can therefore have the same mean-corrected contribution at the next layer while producing very different changes in the logits.

Loss-gradient utilities capture a complementary signal. They can encourage plasticity by favoring replacement of units with weak current learning signals, but they are not reliable as a standalone rank-based reset-cost metric. Let \(J_i(x)=\partial z_\theta(x)/\partial h_i(x)\) be the downstream Jacobian from unit \(i\) to the logits, let \(e(x)=\nabla_z \mathcal L(z_\theta(x),y)\), and let \(\Delta z_i(x)\) denote the linearized reset impact on the logits from replacing \(h_i(x)\) with \(r_i\). The loss gradient and this linearized reset impact are
\begin{equation}
\begin{aligned}
\left|\frac{\partial \mathcal L}{\partial h_i}\right|
&=
\left|e(x)^\top J_i(x)\right|,
\\
\|\Delta z_i(x)\|
&\approx
\left\|J_i(x)(h_i(x)-r_i)\right\|.
\end{aligned}
\label{eq:loss_gradient_reset_impact}
\end{equation}
Both quantities involve downstream sensitivity, but they answer different questions. The loss gradient is gated by the current residual direction \(e(x)\), whereas reset impact is gated by the displacement \(h_i(x)-r_i\). A unit can have small loss gradient because the current example is already fit, or because its output direction is nearly orthogonal to the residual, while replacing it would still perturb the logits. Conversely, a large loss gradient can occur for a unit whose reset displacement is small. Thus, loss-gradient utility can be valuable as a plasticity heuristic, but by itself it can rank as disposable units that still have high functional reset cost.

These cases point to the same missing structure. A reset-cost utility should combine the displacement induced by the reset with the downstream sensitivity of the model output. Activation scores are too local, CBP contribution estimates a one-hop compensated perturbation, and loss-gradient scores emphasize plasticity pressure rather than reset impact. The next section derives a utility that combines reset displacement and downstream output sensitivity directly through reference-based attribution.

\begin{figure}[t]
  \centering
  \includegraphics[width=\linewidth]{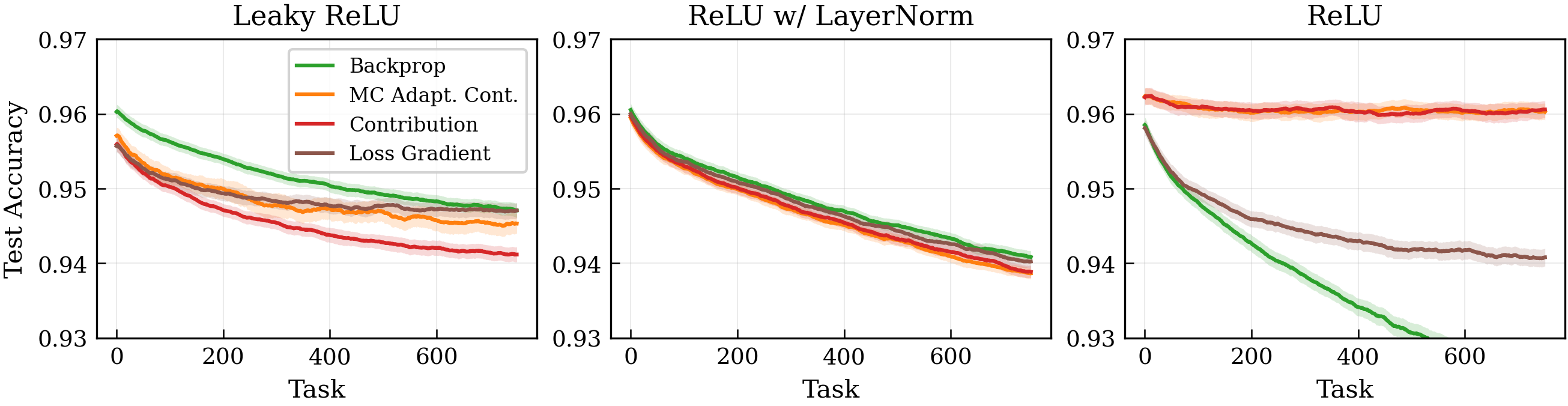}
  \caption{Failure cases of existing utility measures on Online Permuted MNIST. \textbf{Left:} Under Leaky ReLU, existing utilities (defined in Table~\ref{tab:app_utilities}) fail to prevent plasticity loss. \textbf{Center:} In a network with LayerNorm, existing utilities also degrade. \textbf{Right:} In the ReLU feedforward case, Loss Gradient degrades to near-backprop performance. }
  \label{fig:utility_failures}
\end{figure}

\section{Attribution-based neuron utility}

\subsection{From reset cost to reference-based attribution}

Section~3.2 frames reset selection as a tradeoff between future plasticity
benefit and immediate functional cost. We now derive a tractable approximation
to the cost term \(C_t(\{i\})\). Exact evaluation would require applying a
candidate reset to each mature unit and measuring the resulting change in the
network output. Instead, our proposed method approximates this intervention
directly in function space.

This function-space approximation turns reset selection into an attribution
problem. Shapley-style attribution defines a component's contribution by its
average marginal effect over coalitions of present and absent components
\citep{shapley1953value,lundberg2017unified}. This provides a principled notion
of marginal contribution, but it is more general than the intervention used by
CBP, which applies a concrete reset operator to a small set of units in the
current network state. We therefore use Shapley-style attribution as motivation
for a marginal-effect view of utility, while specializing the attribution target
to the reset actually performed by CBP.

Reference-based attribution methods provide the closer analogue. DeepLIFT,
Integrated Gradients, and gradient \(\times\) difference-from-reference attribute
output changes to the displacement of an internal variable from a chosen
baseline \citep{shrikumar2017learning, sundararajan2017axiomatic,
ancona2018towards, shrikumar2018computationally}. CBP's bias-compensated reset
has the same local form: removing a unit while adding back its average
contribution to downstream biases is equivalent to replacing its current
activation \(h_i(x)\) with the running activation estimate used by the reset. We
therefore set \(r_i=r_{l,i,t}=\hat f_{l,i,t}\), making the attribution baseline
the endpoint of the reset intervention rather than an arbitrary zero state.

Let \(\delta_i(x)=h_i(x)-r_i\) denote the reset displacement. Let
\begin{equation}
  J_i(x)
  =
  \frac{\partial z_\theta(x)}{\partial h_i(x)}
  \in \mathbb{R}^C
\end{equation}
be the downstream Jacobian from unit \(i\) to the logits. The first-order logit
perturbation induced by the reset is
\begin{equation}
  \Delta z_i(x)
  \approx
  J_i(x)\delta_i(x).
  \label{eq:linearized_reset_impact}
\end{equation}
The corresponding logit-vector reset-cost proxy is
\begin{equation}
  U_i^{\mathrm{all\text{-}logit}}
  =
  \mathbb{E}_x
  \left[
    \left\|J_i(x)(h_i(x)-r_i)\right\|
  \right].
  \label{eq:all_logit_gxd}
\end{equation}
This expression is the output-impact principle behind GXD: estimate the
functional perturbation caused by moving a unit to the state used by the reset
mechanism.

\begin{figure}[t]
  \centering
  \begin{minipage}[t]{0.32\linewidth}
    \centering
    \includegraphics[width=\linewidth]{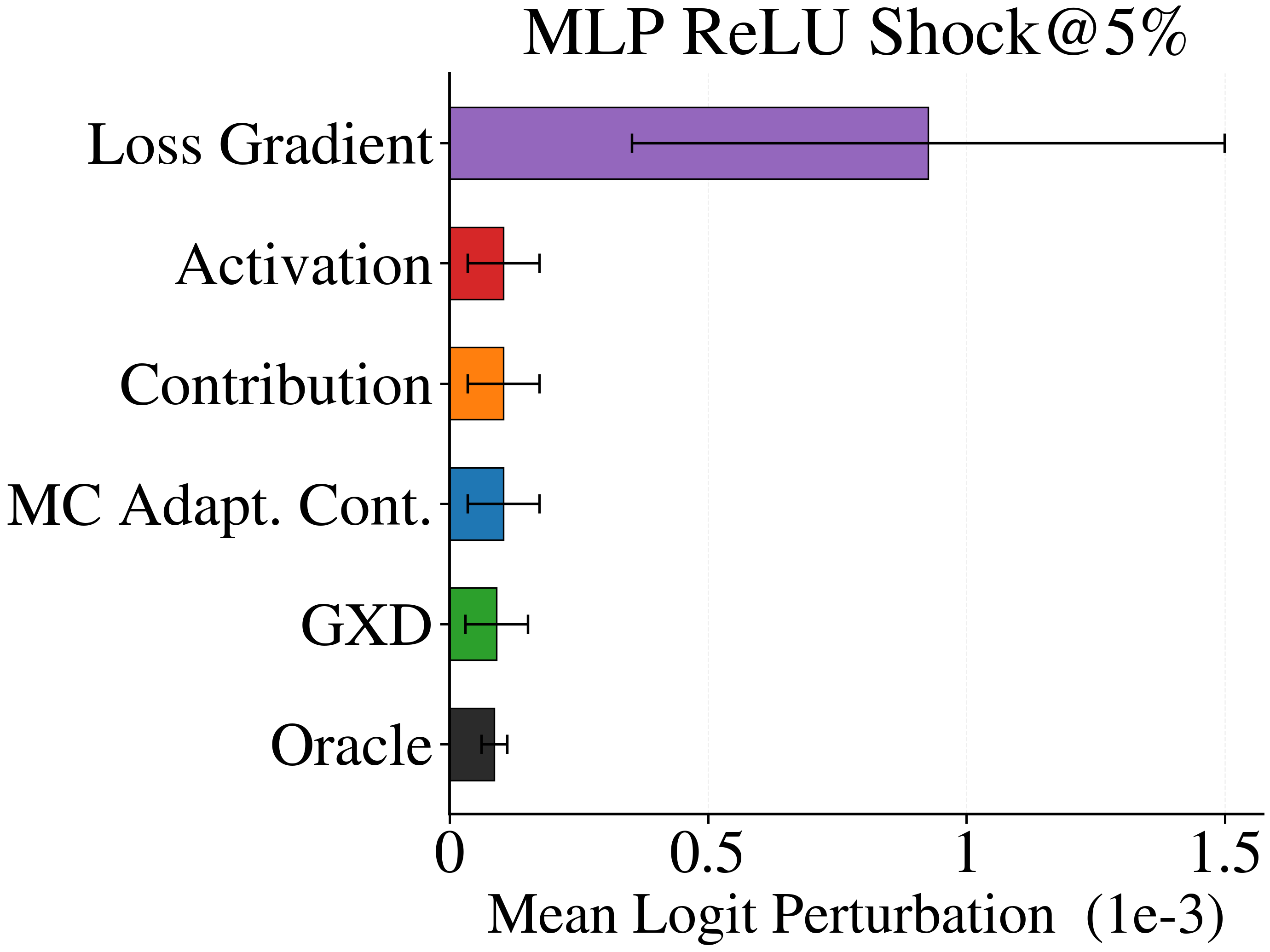}
  \end{minipage}
  \hfill
  \begin{minipage}[t]{0.32\linewidth}
    \centering
    \includegraphics[width=\linewidth]{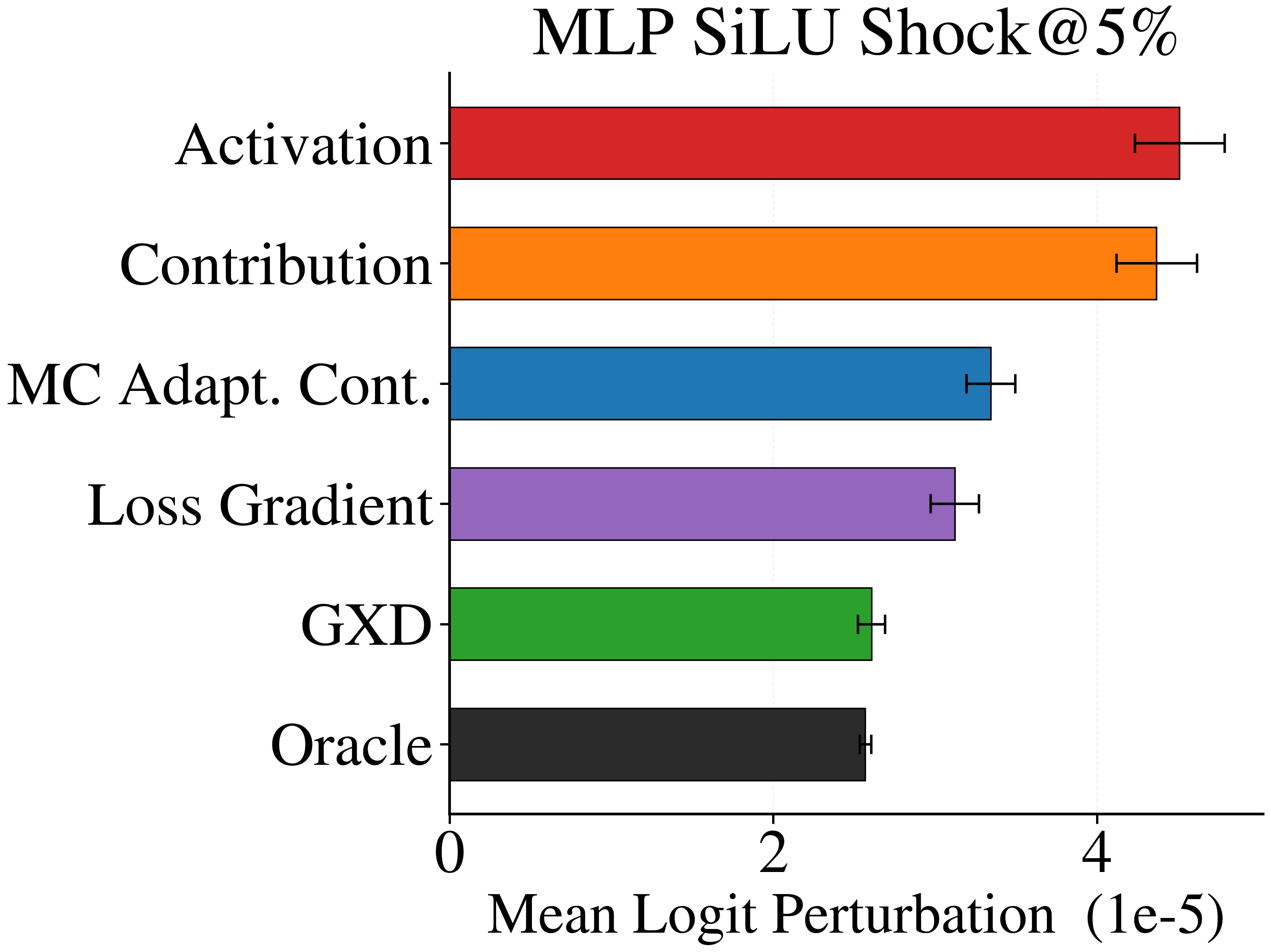}
  \end{minipage}
  \hfill
  \begin{minipage}[t]{0.32\linewidth}
    \centering
    \includegraphics[width=\linewidth]{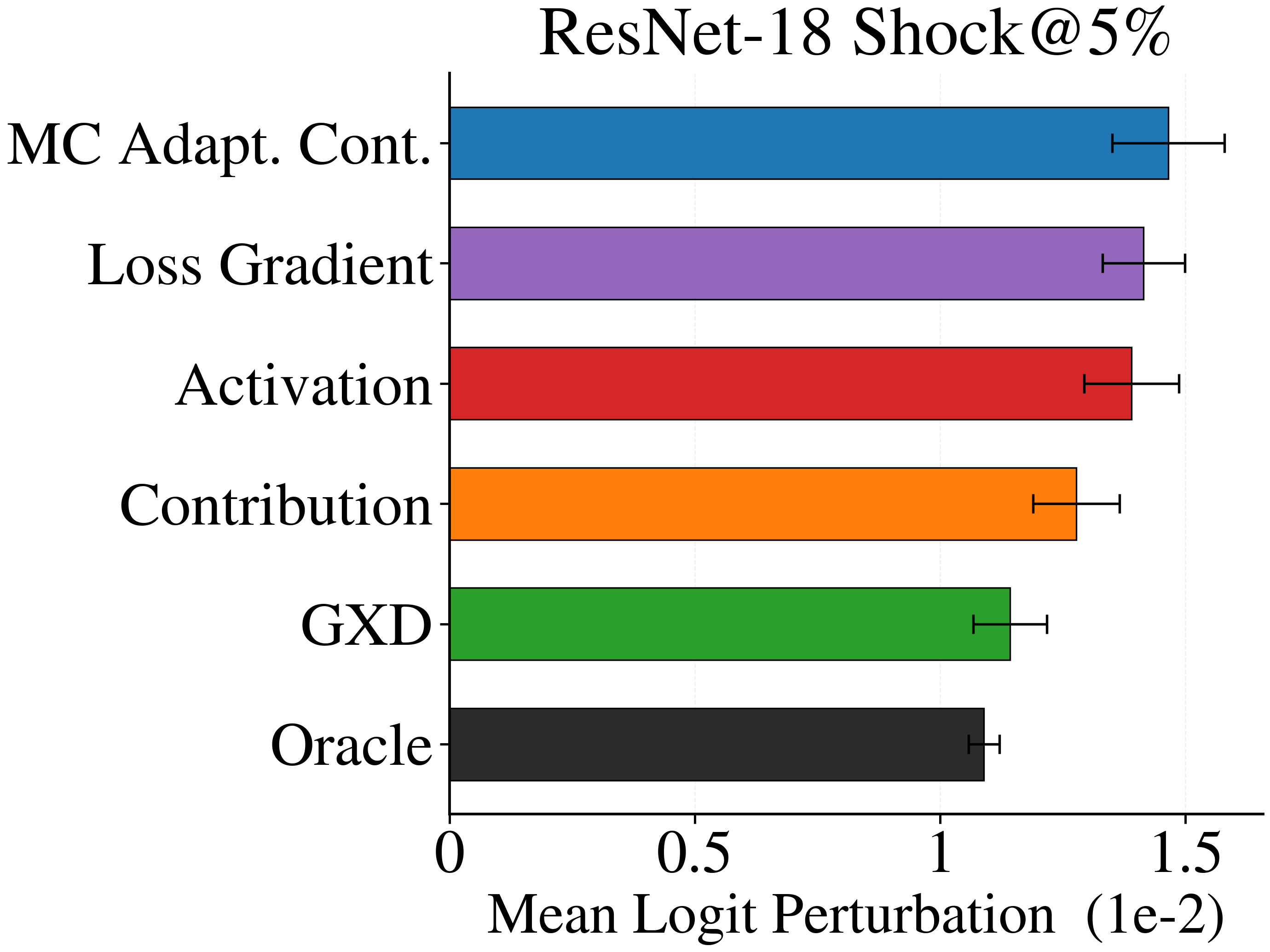}
  \end{minipage}
  \caption{Shock@5\%: mean output perturbation from individually setting each neuron of the bottom 5\% of a utility rank to its reset reference. Lower is better. \textbf{Left:} MLP with ReLU---Loss Gradient has the largest shock. \textbf{Center:} MLP with SiLU---local utilities become unreliable. \textbf{Right:} ResNet-18 (100 heads); GXD consistently produces the lowest reset shock across architectures.}
  \label{fig:shock5}
\end{figure}

\subsection{Target-logit GXD}
\label{sec:target_gxd}

The implementation studied in this paper uses a scalar output functional rather
than the full logit vector. For a scalar score \(S(x)\), the first-order
perturbation is
\begin{equation}
  \Delta S_i(x)
  \approx
  (h_i(x)-r_i)
  \frac{\partial S(x)}{\partial h_i(x)} .
\end{equation}

For supervised classification, we use the target logit,
\(S_y(x)=z_y(x)\). This avoids scaling the attribution computation with the
number of output dimensions while keeping the approximation focused on a
task-relevant output. The target logit is already produced by the ordinary
forward pass, so computing its attribution requires only one additional backward
pass. Averaging the resulting score over datapoints also makes the utility
follow the class balance of the sampled data, avoiding an additional
class-weighting rule. GXD therefore estimates the scalar perturbation as
\begin{equation}
  u_i^{\mathrm{GXD}}(x,y)
  =
  \left|
    (h_i(x)-r_i)
    \frac{\partial z_y(x)}{\partial h_i(x)}
  \right|
  \label{eq:target_gxd}
\end{equation}
This estimates the magnitude of the first-order perturbation to the
correct-class logit caused by moving neuron \(i\) from its current activation to
its reset reference. Appendix~\ref{app:gxd_variants} compares this efficient
target-logit approximation with an all-logit GXD variant and finds nearly
identical reset-cost performance.

The output gradient also connects GXD to the learning signal available to the
unit. As shown in Equation~\ref{eq:loss_gradient_reset_impact}, \(J_i(x)\) is
the downstream sensitivity map through which output-level error signals are
assigned to neuron \(i\). GXD uses this same downstream sensitivity, but weights
it by reset displacement \(h_i(x)-r_i\) rather than by the current loss residual.
This makes output-gradient sensitivity a promising incentive for trainability,
similar in spirit to gradient-based recycling signals such as ReGraMa
\citep{liu2025regrama}. In this paper, however, we validate the combined GXD
score as a reset-cost estimator; whether output-gradient sensitivity is useful as
a standalone trainability signal requires separate evaluation.

\subsection{Batch aggregation and online ranking}

In CBP, the instantaneous utility is converted into a stable online ranking
signal. For each minibatch \(B_t\), we compute the absolute target-logit GXD
score from Equation~\ref{eq:target_gxd} for every example and average those
scores over the minibatch, giving \(\hat{u}_{i,t}^{\mathrm{GXD}}\). Averaging
absolute per-example perturbation estimates avoids cancellation between positive
and negative attributions. We then maintain an EWMA utility estimate,
\begin{equation}
  u_{i,t}
  =
  \gamma u_{i,t-1}
  +
  (1-\gamma)\hat{u}_{i,t}^{\mathrm{GXD}},
\end{equation}
where $\gamma \in [0,1)$ is the utility decay parameter. This running score is used exactly as in standard CBP: after the maturity threshold is reached, the lowest-utility units are selected for replacement under the fixed replacement schedule.

\section{Experiments}

We evaluate GXD across three experimental settings that progressively test different aspects of reset utility quality. First, a controlled lesion study (\S\ref{sec:reset_cost}) directly measures how well each utility predicts the output perturbation caused by a reset, comparing rankings on MLP and ResNet architectures. Second, a Permuted MNIST continual learning benchmark (\S\ref{sec:permuted_mnist}) tests whether better reset-cost estimation translates to sustained plasticity over hundreds of sequential tasks. Third, a Continuous CIFAR-100 experiment (\S\ref{sec:cifar100}) evaluates feature stability in a ResNet under repeated class exposure.

\begin{table}[b]
\centering
\caption{Spearman rank correlation ($\rho$) between each utility score and realized output perturbation (mean $\pm$ SE across checkpoints).  Higher is better; \textbf{bold} marks the best in each column.}
\label{tab:spearman}
\small
\setlength{\tabcolsep}{4pt}
\begin{tabular}{l cc cc cc}
\toprule
 & \multicolumn{2}{c}{MLP ReLU (10 Classes)} & \multicolumn{2}{c}{MLP SiLU (10 Classes)} & \multicolumn{2}{c}{ResNet-18 (100 Classes)} \\
\cmidrule(lr){2-3} \cmidrule(lr){4-5} \cmidrule(lr){6-7}
Utility & Logit \(L_1\) & KL & Logit \(L_1\) & KL & Logit \(L_1\) & KL \\
\midrule
Activation         & $.903 {\scriptstyle\pm .019}$ & $.892 {\scriptstyle\pm .022}$ & $.146 {\scriptstyle\pm .050}$ & $.131 {\scriptstyle\pm .042}$ & $.766 {\scriptstyle\pm .008}$ & $.620 {\scriptstyle\pm .016}$ \\
Contribution       & $.917 {\scriptstyle\pm .019}$ & $.904 {\scriptstyle\pm .022}$ & $.194 {\scriptstyle\pm .041}$ & $.156 {\scriptstyle\pm .037}$ & $.832 {\scriptstyle\pm .006}$ & $.679 {\scriptstyle\pm .016}$ \\
MC Adapt. Cont.    & $.960 {\scriptstyle\pm .013}$ & $.944 {\scriptstyle\pm .016}$ & $.697 {\scriptstyle\pm .056}$ & $.433 {\scriptstyle\pm .051}$ & $.743 {\scriptstyle\pm .011}$ & $.576 {\scriptstyle\pm .019}$ \\
Loss Gradient      & $.719 {\scriptstyle\pm .074}$ & $.733 {\scriptstyle\pm .071}$ & $.726 {\scriptstyle\pm .025}$ & $.446 {\scriptstyle\pm .041}$ & $.751 {\scriptstyle\pm .014}$ & $.680 {\scriptstyle\pm .016}$ \\
GXD                & $\mathbf{.996} {\scriptstyle\pm .001}$ & $\mathbf{.984} {\scriptstyle\pm .003}$ & $\mathbf{.980} {\scriptstyle\pm .002}$ & $\mathbf{.560} {\scriptstyle\pm .051}$ & $\mathbf{.926} {\scriptstyle\pm .003}$ & $\mathbf{.803} {\scriptstyle\pm .012}$ \\
\bottomrule
\end{tabular}
\end{table}

\subsection{Minimizing Reset Cost in MLPs and ResNets}
\label{sec:reset_cost}

We first isolate the ranking problem by measuring reset cost directly at frozen
checkpoints. For each utility, we rank mature hidden units on a calibration set
and then evaluate the same units on a disjoint probe set by clamping one unit at
a time to the reset reference used by CBP, \(h_i(x) \leftarrow r_i\), while
leaving all parameters unchanged. This produces perturbed logits
\(z^{[i\rightarrow r_i]}(x)\), which we compare to the original logits \(z(x)\)
using logit \(L_1\) distance,
\(\|z(x)-z^{[i\rightarrow r_i]}(x)\|_1\), and KL divergence. We summarize low-utility
selection quality with Shock@5\%: the mean perturbation from clamping each of the bottom
5\% of units per layer under each ranking. This perturb-and-measure protocol follows
prior ablation-based evaluations of internal-neuron importance, which clamp units to a
reference value and compare the resulting output changes
\citep{dhamdhere2019important,shrikumar2018computationally}. We measure full-ranking quality with Spearman
correlation between utility score and realized perturbation. We run the procedure on
ReLU and SiLU MLPs trained on Permuted MNIST and on a ResNet-18 architecture \citep{he2016deep}, trained on
CIFAR-100. Across all settings, the reset intervention is fixed and only the
utility ranking changes.

\paragraph{Results}
GXD is the most consistent reset-cost estimator across the architectures and
activation functions tested. In the ReLU MLP, where local contribution scores are
already strong, GXD nearly saturates the rank correlation with measured logit
\(L_1\) shock. The separation is larger in the SiLU MLP: activation and CBP
contribution correlate weakly with realized perturbation, while GXD remains
highly correlated. Figure~\ref{fig:shock5} shows the practical consequence of
this ranking difference: the bottom 5\% of units selected by GXD produce much
smaller output shock than those selected by activation-based utilities. In the
ResNet-18, GXD is the best in Spearman correlation and yields the closest Shock@5\% to the oracle (minimum possible impact).
These results show that GXD further minimizes reset perturbation precisely where existing utilities falter, in non-ReLU-activation networks and residual architectures whose skip connections decouple local activity from downstream impact.

\subsection{Maintaining Plasticity on Online Permuted MNIST}
\label{sec:permuted_mnist}

We evaluate GXD within the Continual Backpropagation (CBP) framework on 800 sequential Permuted MNIST tasks, comparing utility estimators while keeping the reset mechanism fixed. Online Permuted MNIST constructs a sequence of classification tasks by applying a new fixed random permutation to the pixels of MNIST \citep{lecun1998gradient} images for each task. This benchmark has been used in recent plasticity studies to expose degradation in a network's ability to adapt over long non-stationary training streams \citep{dohare2024loss,kumar2023maintaining}.

The model is a fully connected network trained for one epoch per task, and we report test accuracy on each task as a measure of maintained plasticity. To test whether reset utility remains reliable when local activity is less directly tied to functional importance, we evaluate ReLU, SiLU, Leaky ReLU, and tanh activations. Results are averaged over 15 random seeds.

\begin{figure}[t]
  \centering
  \includegraphics[width=\linewidth]{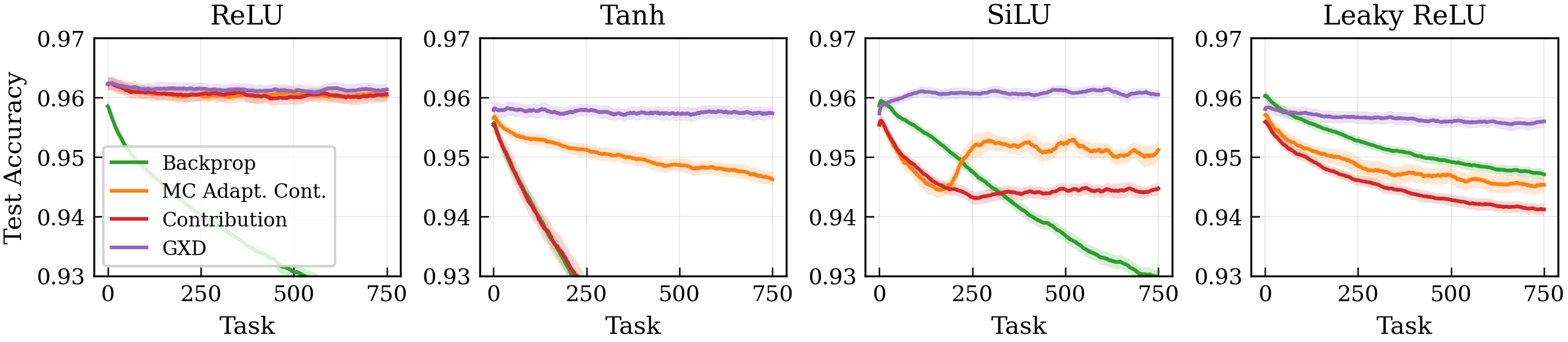}
  \caption{Test accuracy on Permuted MNIST across activation functions. GXD maintains plasticity across ReLU, Tanh, SiLU, and Leaky ReLU, while original CBP's MC Adaptable Contribution and plain Contribution face degradation or instability for non-ReLU activations. Averaged over 15 seeds.}
  \label{fig:permuted_mnist_act}
\end{figure}

\paragraph{Results}

Figure~\ref{fig:permuted_mnist_act} shows that GXD maintains plasticity more reliably than the local utility baselines across the activation regimes tested. In the standard ReLU setting, CBP's original contribution-based utility is already well matched to the activation geometry, and GXD performs comparably. The advantage of GXD becomes clear for SiLU and Leaky ReLU, where local activation levels are less reliable estimates of reset cost, as shown in \S\ref{sec:reset_cost}. In these settings, both contribution utilities degrade over the task sequence, while GXD continues to sustain high accuracy. Tanh provides a complementary case where units can saturate at either positive or negative values, and subtracting a running reference helps correct the offset problem faced by plain activation contribution. As a result, MC Adaptable Contribution degrades less severely than the original contribution score. GXD still performs best, indicating that reference correction is a key improvement, but that the strongest reset criterion also accounts for output sensitivity.

\begin{figure}[t]
  \centering
  \includegraphics[height=1.7in]{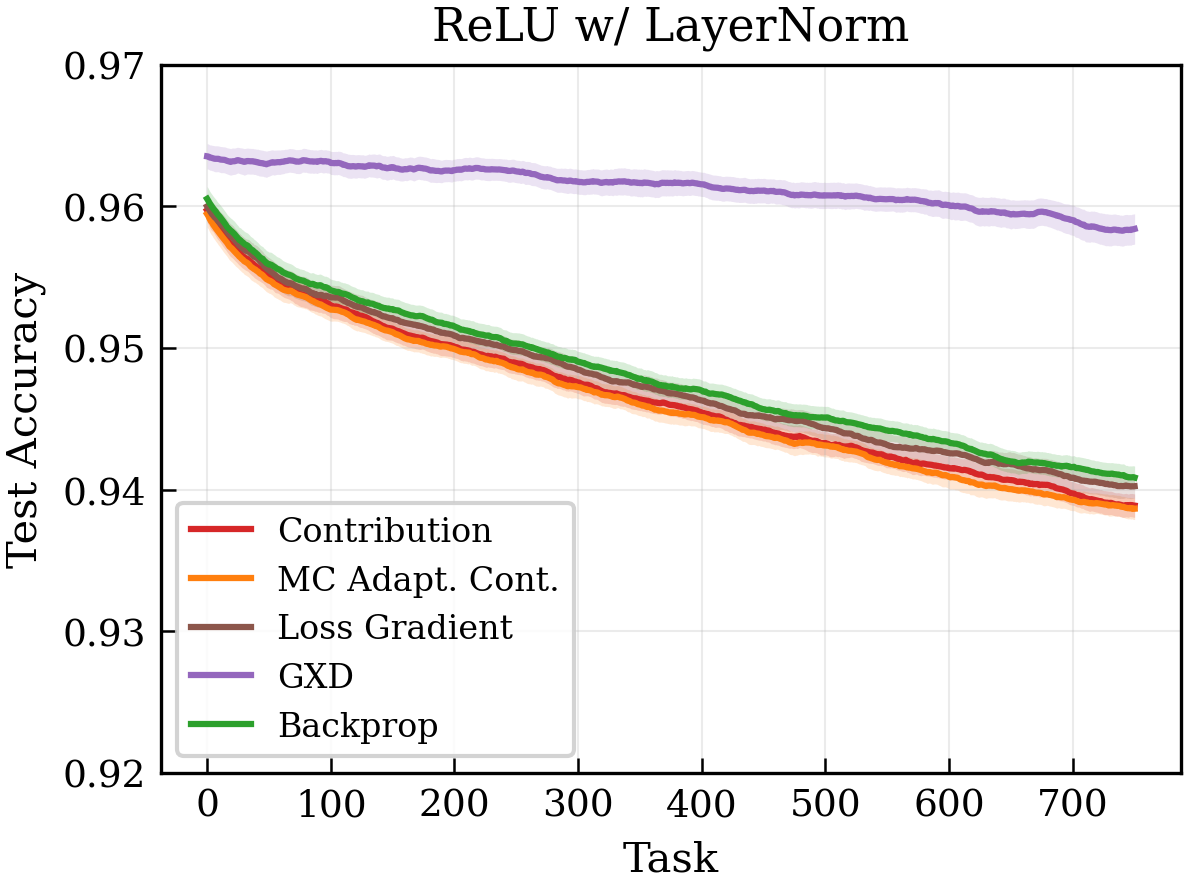}
  \hspace{1.5em}
  \vrule width 0.4pt height 1.7in depth 0pt
  \hspace{1.5em}
  \includegraphics[height=1.7in]{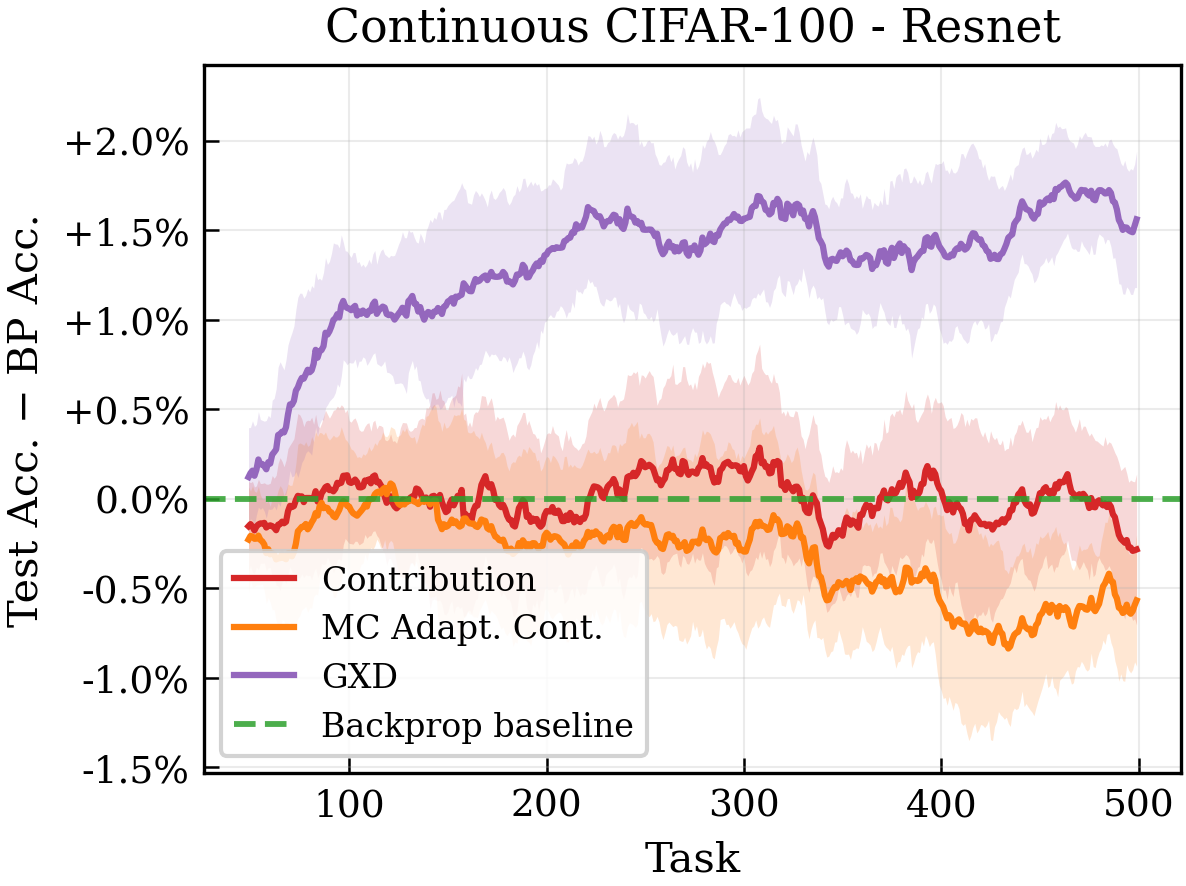}
  \caption{\textbf{Left:} Test accuracy on Permuted MNIST with ReLU and LayerNorm. GXD is the only tested utility that mitigates plasticity loss. \textbf{Right:} Continuous CIFAR-100 with a ResNet. Test accuracy relative to backprop baseline (50-task MA). GXD consistently outperforms Contribution and MC Adaptable Contribution in this feature stability test.}
  \label{fig:layernorm_cifar}
\end{figure}

We also evaluate a ReLU network with layer normalization, shown in Figure~\ref{fig:layernorm_cifar}. All local utility baselines collapse toward standard backpropagation, while GXD is the only tested utility that substantially mitigates plasticity loss. Layer normalization couples units through shared centering and scaling statistics, making one-hop or activation-only scores poor proxies for the cost of replacing a feature. Together, these results show that when local utilities misestimate reset cost, GXD makes CBP's fixed plasticity injection more useful, preserving learned features while selecting low-impact units for replacement, allowing subsequent training to build new capacity rather than repair reset-induced damage.

\subsection{Maintaining Stability on Continuous CIFAR-100}
\label{sec:cifar100}

We next evaluate GXD in a ResNet architecture using Continuous CIFAR-100 \citep{krizhevsky2009learning}, an adaptation of the Continual ImageNet protocol introduced by \citet{dohare2024loss}. Continual ImageNet trains a binary classifier on a sequence of two-class tasks sampled from 1000 classes, allowing hundreds of tasks before classes repeat. CIFAR-100 has only 100 classes, so this adaptation necessarily revisits classes later in training. The experiment is therefore primarily a test of \emph{feature stability}: the model must adapt to each new binary task while preserving features that may be useful when a class reappears.

We design the experiment so that this stability pressure falls on the shared representation rather than the task-specific classifier. The model is a small 8-layer residual network, trained under a fixed stability-test protocol: learning rate 0.1, 200 epochs per task, and replacement rate $\rho=10^{-4}$ for all reset methods. As in \citet{dohare2024loss}, the binary head is reset at each task boundary and only the feature extractor persists task to task. Thus, the utility ranking is the central intervention, as all reset methods use the same nonzero replacement budget and reset mechanism, but choose different feature units to preserve or replace.

Standard backpropagation does not exhibit the monotonic plasticity loss seen in other related experimental settings as shown in Figure~\ref{fig:app_cifar_abs}. This is because binary CIFAR-100 tasks are relatively small for this architecture, and repeated exposure lets the shared feature extractor learn representations for most classes early in training. This absence of collapse makes the benchmark a stability stress test for rank-based resets because the head is reinitialized at every task boundary, performance on a revisited class depends on whether the persistent feature extractor has retained useful class structure.

\paragraph{Results}

GXD performs best in this stability setting (Figure~\ref{fig:layernorm_cifar}, right). Relative to the backpropagation baseline, it maintains higher accuracy and exhibits smaller performance drawdowns than Contribution and MC Adaptable Contribution. This is consistent with the reset-cost view: GXD ranks eligible units by an estimate of the functional perturbation induced by resetting them, so its replacements are biased toward units whose removal least changes the current predictor. On Continuous CIFAR-100, that stability advantage translates into better reuse of previously learned features when classes recur.

\section{Discussion}
GXD improves the reset-selection component of Continual Backpropagation by estimating the functional cost of resetting a unit, aligned with the reset mechanism itself. This is useful when functional importance is obscured by non-ReLU activations, normalization, or skip connections. Empirically, GXD better predicts reset-induced output changes and improves Continual Backpropagation performance in the settings tested. Future work will extend this intervention-aligned view of utility to more complex environments such as transformers and reinforcement learning. These settings may also require moving beyond neuron-level resets toward weight-level resets \citep{hernandez2025selective} or partial resets \citep{mccutcheon2026continuous}, with attribution methods aligned to the corresponding reset operation.

\paragraph{Limitations.}
Our experiments are limited to supervised continual learning with MLPs with activation and normalization variants, convolutional layers, and small ResNets. We have not yet evaluated GXD on attention, LLMs, large-scale vision models, or RL agents. Our experiments are intentionally an isolated study of the utility measure within rank-based neuron resets, and future work should compare CBP-GXD to more recent state-of-the-art plasticity mitigation methods. GXD is also a first-order, per-unit approximation which does not account for nonlinear curvature in the effect of a reset and does not estimate the combined effect of resetting multiple units at once. Future work could explore higher-order or set-aware reset-cost estimators that account for these effects. In addition, the current implementation uses an efficient scalar target-logit attribution, which may need to be extended for architectures with a large number of output heads. Finally, our experiments validate GXD primarily as a reset-cost estimator and future work could empirically evaluate output-gradient sensitivity as a standalone signal for trainability.

\section*{Acknowledgements}        
The views expressed here are those of the authors alone and not of The Vanguard Group, Inc.

{\small

\bibliographystyle{plainnat}

}

\appendix

\section{Experimental Details}
\label{app:experimental_details}

Table~\ref{tab:app_utilities} summarizes all utility methods compared across experiments. All use EMA-tracked statistics with decay $\beta = 0.99$ and bias correction $u_i / (1 - \beta^{a_i})$ where applicable. GXD variants are detailed in Appendix~\ref{app:gxd_variants}.

\begin{table}[h]
\centering
\caption{Utility methods. $h_i$: post-activation; $r_i$: bias-corrected EMA reference; $w^{\mathrm{out}}, w^{\mathrm{in}}$: outgoing/incoming weights; $z_y$: target logit; $\mathcal{L}$: loss.}
\label{tab:app_utilities}
\small
\setlength{\tabcolsep}{4pt}
\begin{tabular}{p{0.26\linewidth}p{0.66\linewidth}}
\toprule
\textbf{Utility} & \textbf{Formula} \\
\midrule
Activation & $\mathbb{E}[\lvert h_i \rvert]$ \\
Contribution & $\mathbb{E}[\lvert h_i \rvert] \cdot \sum_k \lvert w_{i,k}^{\mathrm{out}} \rvert$ \\
MC Adapt. Cont. & $\lvert h_i - r_i \rvert \cdot \sum_k \lvert w_{i,k}^{\mathrm{out}} \rvert \;/\; \sum_j \lvert w_{j,i}^{\mathrm{in}} \rvert$ \\
Loss Gradient & $\mathbb{E}[\lvert \partial \mathcal{L} / \partial h_i \rvert]$ \\
GXD & $\mathbb{E}[\lvert (h_i - r_i) \cdot \partial z_y / \partial h_i \rvert]$ \\
\bottomrule
\end{tabular}
\end{table}

\subsection{Continual Backpropagation Details}
\label{app:cbp_details}

For hidden unit \(i\) in layer \(l\) at update \(t\), let \(h_{l,i,t}(x)\) be the post-activation value and \(a_{i,t}^{(l)}\) be the unit age. CBP utilities maintain recent activation statistics. The signed running activation reference used by mean-corrected utilities is
\begin{equation}
  f_{l,i,t} = (1-\eta) h_{l,i,t} + \eta f_{l,i,t-1},
  \qquad
  r_{l,i,t}=\hat f_{l,i,t}=\frac{f_{l,i,t}}{1-\eta^{a_{i,t}^{(l)}}},
\end{equation}
where the bias correction removes the zero-initialization bias of the EMA, making \(r_{l,i,t}\) an age-corrected estimate of the unit's recent average activation.

Contribution utility estimates local expression weighted by outgoing connection magnitude. It is calculated as an EMA of instantaneous outgoing contribution,
\begin{equation}
  u_{l,i,t}^{\mathrm{Cont}}
  =
  \eta u_{l,i,t-1}^{\mathrm{Cont}}
  +
  (1-\eta)
  \left|h_{l,i,t}\right|
  \sum_{k=1}^{n_{l+1}}\left|w_{i,k,t}^{(l)}\right|,
\end{equation}
with \(\eta=0.99\) in \citet{dohare2024loss}. Mean-corrected contribution replaces \(\lvert h_{l,i,t}\rvert\) with displacement from the signed reference, \(\lvert h_{l,i,t}-r_{l,i,t}\rvert\). Mean-corrected adaptable contribution, denoted as the overall contribution in this version: \citep{dohare2024maintaining}, further divides by incoming weight magnitude,
\begin{equation}
  y_{l,i,t}^{\mathrm{MCAdapt}}
  =
  \frac{
    \left|h_{l,i,t}-r_{l,i,t}\right|
    \sum_{k=1}^{n_{l+1}} \left|w_{i,k,t}^{(l)}\right|
  }{
    \sum_{j=1}^{n_{l-1}} \left|w_{j,i,t}^{(l-1)}\right|
  }.
\end{equation}
The tracked utilities are an EMA of the instantaneous score, and mature units are ranked by its bias-corrected value. 

\begin{algorithm}[H]
\caption{Continual Backpropagation for an MLP}
\label{alg:cbp}
\begin{algorithmic}[1]
\Require Replacement rate \(\rho\), decay rate \(\eta\), maturity threshold \(m\)
\State Initialize weights with \(w\sim d_l\), where \(d_l\) is the initialization distribution
\State Initialize utilities \(u\), replacement counters \(c\), and ages \(a\) to zero
\For{each input \(x_t\)}
  \State Train step
  \For{each hidden layer \(l=1,\ldots,L-1\)}
    \State Update ages: \(a_l\leftarrow a_l+1\)
    \State Update utilities \(u_l\) with decay rate \(\eta\)
    \State Count mature units: \(n_{\mathrm{eligible}}\leftarrow |\{i:a_{l,i}>m\}|\)
    \State Accumulate replacements: \(c_l\leftarrow c_l+n_{\mathrm{eligible}}\rho\)
    \If{\(c_l>1\)}
      \State Select \(r=\arg\min_{i:a_{l,i}>m} u_{l,i}\)
      \State Reinitialize incoming weights: resample \(w_{l-1}[:,r]\sim d_l\)
      \State Compensate downstream bias: \(b_l\leftarrow b_l+w_l[r,:]\hat f_{l,r}\)
      \State Reinitialize outgoing weights: set \(w_l[r,:]\leftarrow 0\)
      \State Reset utility and age: \(u_{l,r}\leftarrow 0\), \(a_{l,r}\leftarrow 0\)
      \State Update replacement counter: \(c_l\leftarrow c_l-1\)
    \EndIf
  \EndFor
\EndFor
\end{algorithmic}
\end{algorithm}

\subsection{Reset Cost Assay (\S\ref{sec:reset_cost})}
\label{app:reset_cost}

Models are trained and checkpointed at fixed intervals. For each checkpoint, utility scores are computed on a calibration set (2{,}048 samples), and the output perturbation from clamping low-utility neurons to their reference, \(r_i \approx \mathbb{E}[h_i(x)]\), is measured on a disjoint probe set (2{,}048 samples). We measure output shock using logit \(L_1\) distance,
and KL divergence between the original and perturbed model
outputs. \(L_1\) measures absolute functional
displacement, while KL measures effective change in the predictive
distribution. Similar output-distance criteria have been used to quantify
changes in model predictions \citep{luo2020neural,xu2018feature}. The assay additionally evaluates GXD variants listed in Appendix~\ref{app:gxd_variants}.

MLP metrics: mean $\pm$ SE across 3 seeds $\times$ 5 checkpoints. ResNet-18 metrics: 3 seeds $\times$ 4 checkpoints.
For all ResNet architectures, a BasicBlock denotes the residual block of \citet{he2016deep}: two 3$\times$3 convolution--BatchNorm--ReLU layers with an identity shortcut when dimensions match and a projection shortcut when channels or spatial resolution change.

\begin{table}[H]
\centering
\caption{Reset Cost Assay: architecture and training.}
\label{tab:app_reset_cost}
\small
\setlength{\tabcolsep}{4pt}
\begin{tabular}{p{0.20\linewidth}p{0.30\linewidth}p{0.42\linewidth}}
\toprule
& \textbf{MLP (Permuted MNIST)} & \textbf{ResNet-18 (CIFAR-100)} \\
\midrule
Architecture & 784$\to$[256]$\times$4$\to$10 & ResNet-18 \citet{he2016deep}, also used in \citet{dohare2024loss}: 4 stages with [2,2,2,2] BasicBlocks; [64, 128, 256, 512] channels \\
Activations & ReLU, SiLU, LReLU, Tanh & ReLU \\
Normalization & None & BatchNorm2d \\
Init & Kaiming uniform & Kaiming normal \\
\midrule
Optimizer & SGD (lr=0.01) & SGD (lr=0.1, mom=0.9, wd=$5{\times}10^{-4}$) \\
LR schedule & None & $\times$0.2 at epochs 60, 120, 160 \\
Batch size & 64 & 90 / 100 \\
Tasks / epochs & 30 tasks, 1 epoch each & 200 epochs, all 100 classes \\
Data split & Full train / test & 450 train + 50 val per class \\
Augmentation & None & Flip, crop(pad=4), rot($0$--$15^\circ$) \\
Seeds & 3 & 3 \\
Checkpoints & Tasks \{0, 5, 10, 20, 30\} & Epochs \{60, 120, 160, 200\} \\
\midrule
Compute & CPU & NVIDIA A10G (24\,GB) \\
\bottomrule
\end{tabular}
\end{table}

\subsection{Online Permuted MNIST (\S\ref{sec:permuted_mnist})}
\label{app:permuted_mnist}

\begin{table}[H]
\centering
\caption{Permuted MNIST: configuration.}
\label{tab:app_pmnist}
\small
\begin{tabular}{p{0.28\linewidth}p{0.64\linewidth}}
\toprule
\textbf{Parameter} & \textbf{Value} \\
\midrule
Architecture & FC 784$\to$[256]$\times$4$\to$10, init: Glorot uniform \\
Activations & ReLU, SiLU, Leaky ReLU, Tanh \\
Normalization & None (LayerNorm tested seperately) \\
Tasks & 800 sequential MNIST permutations \\
Epochs / task & 1 \\
Batch size & 16 \\
Optimizer & SGD (no momentum) \\
\midrule
LR grid & \{0.1, 0.3\} \\
Replacement rate grid & \{$10^{-5}$, $10^{-4}$, $10^{-3}$\} (backprop uses 0) \\
Selection & Best final accuracy of LR $\times$ RR per method--activation \\
Selected LR & Backprop: 0.1 for all activations; CBP methods: 0.3 for all activations \\
Selected RR & Backprop: 0; CBP methods: $10^{-4}$ for ReLU/Tanh, $10^{-3}$ for SiLU/Leaky ReLU \\
\midrule
Methods & Backprop, Contribution, MC Adapt. Contr., GXD, Loss Gradient \\
Seeds & 15 \\
Compute & CPU; ${\sim}$4\,h (backprop), ${\sim}$6\,h (Original CBP), ${\sim}$6.1\,h (GXD CBP) per seed \\
\bottomrule
\end{tabular}
\end{table}

MNIST pixels normalized to $[0, 1]$; permutations generated from a fixed seed shared across methods. Hyperparameters were selected from 5-seed sweeps before running the final reported seeds. All reset-based methods received the same learning-rate and replacement-rate candidate set. Hyperparameters not explicitly swept or fixed above follow the prior CBP/plasticity-loss settings used for Permuted MNIST \citep{dohare2024loss}. Error bars: SE across 15 seeds. GXD's additional backward pass for the target-logit gradient adds minimal overhead on CPU compared to the original CBP  (${\sim}$6.1\,h vs.\ ${\sim}$6\,h per seed for Activation CBP).

\subsection{Continuous CIFAR-100 (\S\ref{sec:cifar100})}
\label{app:cifar100}

Follows the Continual ImageNet protocol of \citet{dohare2024loss} adapted for CIFAR-100: the model trains on a sequence of 500 binary tasks (random class pairs). The 2-unit head is zeroed and optimizer state is fully reset between tasks; conv layer parameters carry over. Loss gradient utility was not included in this experiment due to is poor performance in Online Permuted MNIST.

\begin{table}[H]
\centering
\caption{Continuous CIFAR-100: architecture and training.}
\label{tab:app_cifar}
\small
\setlength{\tabcolsep}{4pt}
\begin{tabular}{p{0.26\linewidth}p{0.66\linewidth}}
\toprule
\textbf{Parameter} & \textbf{Value} \\
\midrule
Architecture & Small ResNet \citep{he2016deep}: 3$\times$3 conv(16); BasicBlock(32); BasicBlock(64); GAP; 2-unit head \\
BasicBlock & Two 3$\times$3 conv--BatchNorm--ReLU layers plus identity/projection shortcut \\
Normalization & BatchNorm (momentum=0.9) \\
Init & He normal \\
\midrule
Tasks & 500 binary (first 50 exhaust all classes; then random pairs, \\
 & classes may recur but pairings do not repeat) \\
Epochs / task & 200 (fixed stability-test protocol) \\
Optimizer & SGD (momentum=0.9) \\
Weight decay & 0 \\
Batch size & 100 \\
Augmentation & None \\
\midrule
LR grid & \{0.1, 0.3\} \\
Fixed stability-test settings & LR 0.1, 200 epochs/task, $\rho=10^{-4}$ for all reset methods \\
Selection & LR selected from 3 seed sweeps; RR fixed by stability-test design \\
Maturity threshold & 1{,}000 updates \\
\midrule
Methods & Backprop, Contribution, MC Adapt. Contr., GXD\\
Seeds & 10 \\
Compute & NVIDIA A10G (24\,GB), 1 GPU per run \\
 & ${\sim}$1.1\,h (backprop), ${\sim}$1.8\,h (Activation CBP), ${\sim}$1.9\,h (GXD CBP) per seed \\
\bottomrule
\end{tabular}
\end{table}

The Continuous CIFAR-100 experiment is designed as a fixed-budget stability test. We report all reset methods at replacement rate $\rho=10^{-4}$ to compare how different utility rankings allocate the same amount of feature turnover. Hyperparameters not explicitly swept or fixed above follow the Continual ImageNet protocol of \citet{dohare2024loss}. Error bars: SE across 10 seeds.

We use the same layerwise notion of a reset unit as \citet{dohare2024loss}: in convolutional layers, one unit corresponds to an output channel/filter, so resetting a unit reinitializes the incoming filter weights and removes its outgoing connections. In linear layers, one unit corresponds to one hidden feature vector.

\section{GXD Variants}
\label{app:gxd_variants}

The reset cost experiment (\S\ref{sec:reset_cost}) compares the target-logit GXD used in the main experiments against a full all-logit version and GXI (Gradient $\times$ Input), a related attribution baseline. GXD computes $f(h_i - r_i, \nabla_i)$ where $r_i$ is the bias-corrected EMA reference and $\nabla_i$ denotes a gradient with respect to the target logit or, in the all-logit case, averaged over all $C$ output logits. GXI uses the same gradient but replaces the displacement with the raw activation $h_i$, equivalent to a zero reference.

\begin{table}[H]
\centering
\caption{Variants evaluated in the Reset Cost Experiment.}
\label{tab:app_gxd_variants}
\small
\setlength{\tabcolsep}{4pt}
\begin{tabular}{ll}
\toprule
\textbf{Variant} & \textbf{Formula} \\
\midrule
GXI (target) & $\mathbb{E}[\lvert h_i \cdot \partial z_y / \partial h_i \rvert]$ \quad (Gradient $\times$ Input) \\
GXD (target) & $\mathbb{E}[\lvert (h_i - r_i) \cdot \partial z_y / \partial h_i \rvert]$ \\
GXD (all-logit) & $\frac{1}{C}\sum_{c} \mathbb{E}[\lvert (h_i - r_i) \cdot \partial z_c / \partial h_i \rvert]$ \\
\bottomrule
\end{tabular}
\end{table}

\begin{figure}[H]
  \centering
  \begin{minipage}[t]{0.32\linewidth}
    \centering
    \includegraphics[width=\linewidth]{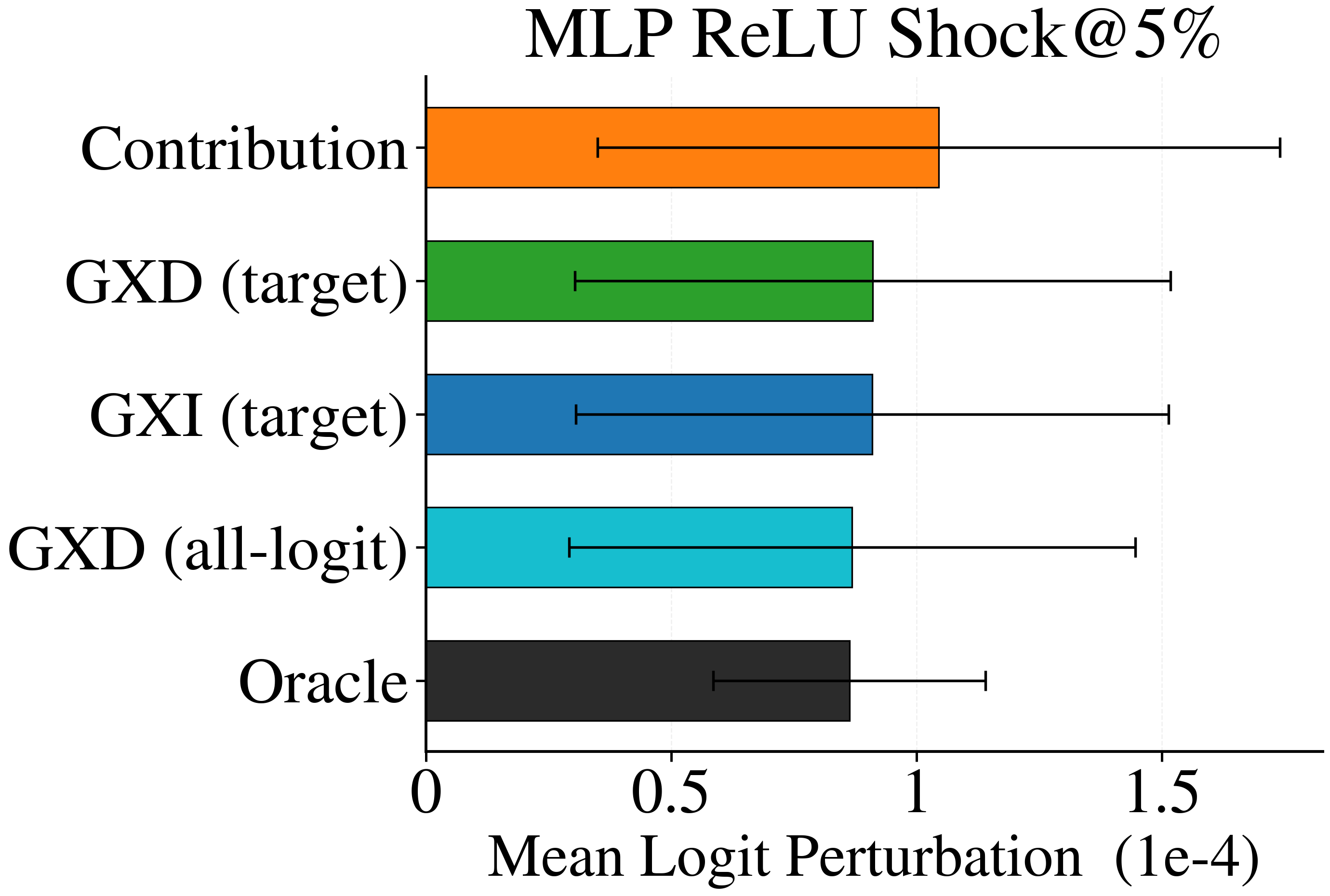}
  \end{minipage}
  \hfill
  \begin{minipage}[t]{0.32\linewidth}
    \centering
    \includegraphics[width=\linewidth]{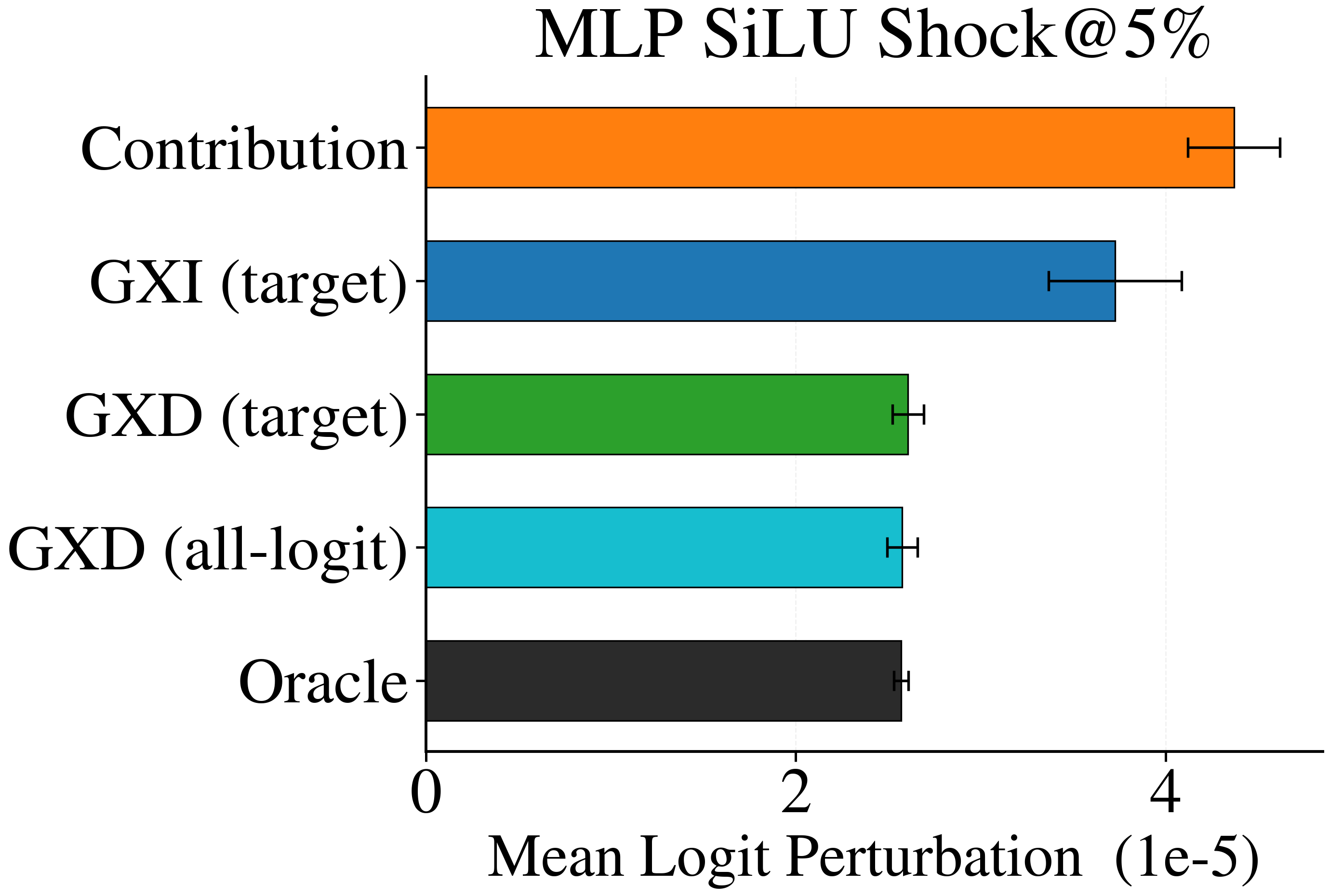}
  \end{minipage}
  \hfill
  \begin{minipage}[t]{0.32\linewidth}
    \centering
    \includegraphics[width=\linewidth]{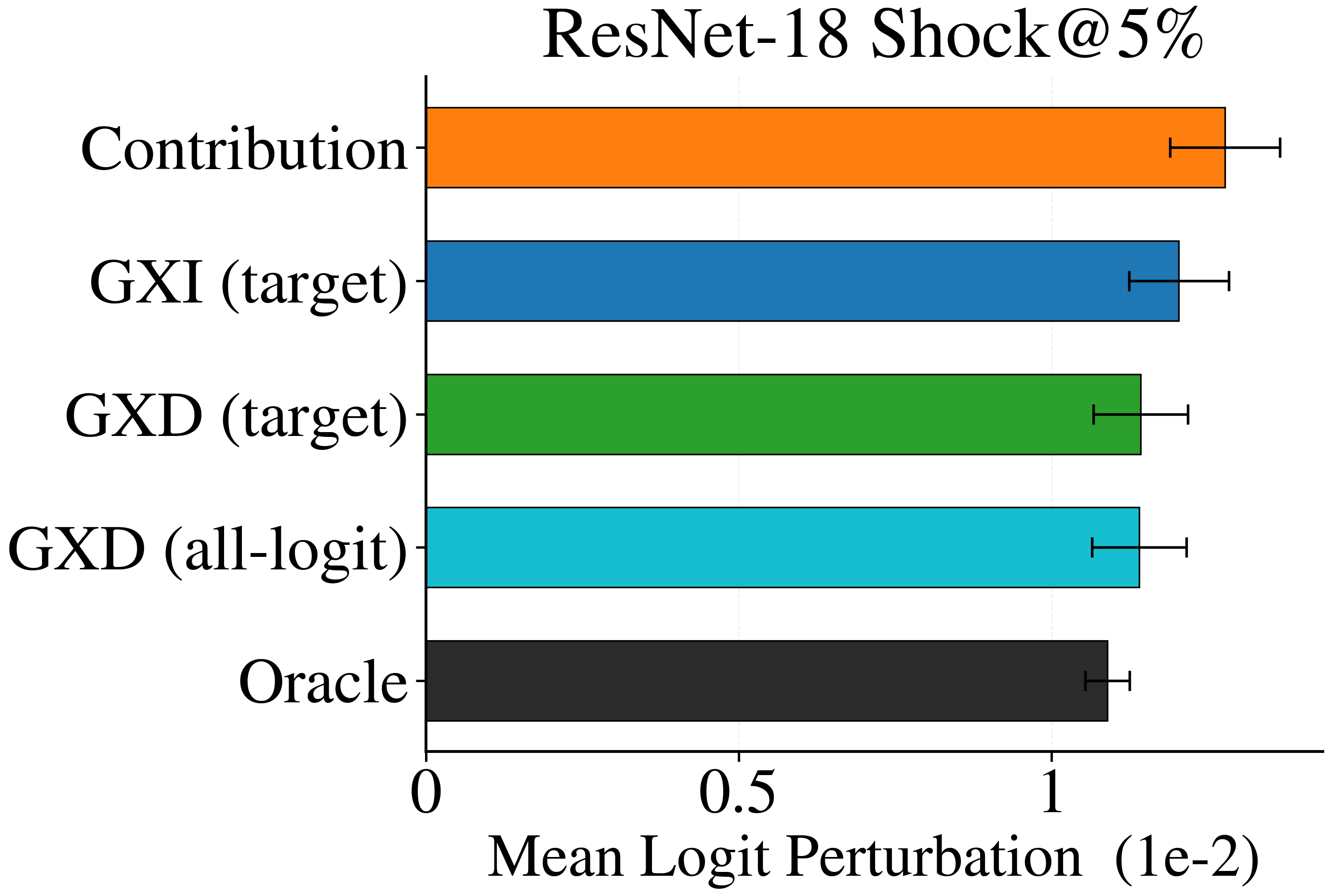}
  \end{minipage}
  \caption{Shock@5\% for GXD variants: GXI (target), GXD (target), and GXD (full-L1). \textbf{Left:} MLP with ReLU. \textbf{Center:} MLP with SiLU. \textbf{Right:} ResNet-18.}
  \label{fig:app_shock5_variants}
\end{figure}

Section~\ref{sec:target_gxd} motivates the target-logit formulation for its efficiency as it only requires one additional backward pass regardless of the number of output classes. Table~\ref{tab:app_spearman_variants} confirms that the ranking quality of GXD (target) nearly matches GXD (all-logit) across all three settings. The gap is especially small for the ResNet-18 (C=100), where Spearman $\rho$ on logit $L_1$ is $.926$ vs.\ $.937$, and GXD (target) actually achieves the higher correlation on KL ($.803$ vs.\ $.783$). Figure~\ref{fig:app_shock5_variants} further shows the alignment of target-logt GXD with the all-logit version where the resetting of the bottom 5\% of each method results in very close resulting perturbation, which is also the minimum among all estimation methods.

\begin{figure}[H]
  \centering
  \includegraphics[width=\linewidth]{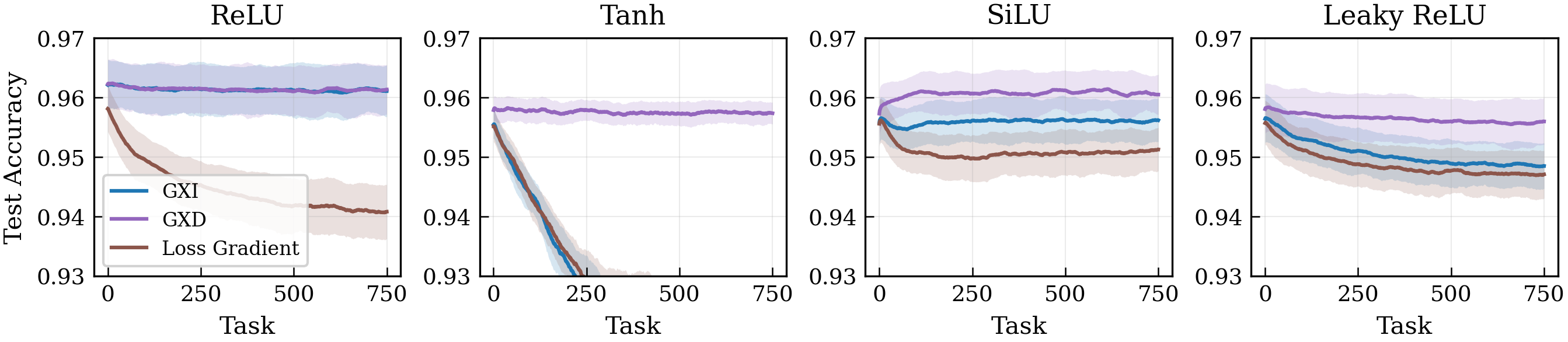}
  \caption{Test accuracy on Permuted MNIST comparing GXI (target), GXD (target), and Loss Gradient utilities across activation functions. Results averaged over 15 seeds.}
  \label{fig:app_pmnist_gxi_grad}
\end{figure}

GXI (Gradient $\times$ Input) is a standard feature attribution method \citep{ancona2018towards} that multiplies a unit's activation by its output gradient. In the GXD framework, GXI corresponds to setting the reference to zero ($r_i = 0$). For ReLU networks, zero is the natural inactive state, so GXI's implicit reference coincides with the reset endpoint. This explains its competitive performance in the ReLU MLP (Table~\ref{tab:app_spearman_variants}: $.970$ vs.\ $.996$ Spearman on logit $L_1$; Figure~\ref{fig:app_shock5_variants} left: comparable Shock@5\% bars; Figure~\ref{fig:app_pmnist_gxi_grad} matches performance of GXD on ReLU).

\begin{table}[H]
\centering
\caption{Spearman rank correlation ($\rho$) for GXD variants (mean $\pm$ SE across checkpoints). GXD (target) is the scalar target-logit formulation used in the main experiments; GXI uses zero as the reference; GXD (all-logit) averages attribution over all output logits.}
\label{tab:app_spearman_variants}
\small
\setlength{\tabcolsep}{4pt}
\begin{tabular}{l cc cc cc}
\toprule
 & \multicolumn{2}{c}{MLP ReLU} & \multicolumn{2}{c}{MLP SiLU} & \multicolumn{2}{c}{ResNet-18} \\
\cmidrule(lr){2-3} \cmidrule(lr){4-5} \cmidrule(lr){6-7}
Utility & Logit \(L_1\) & KL & Logit \(L_1\) & KL & Logit \(L_1\) & KL \\
\midrule
GXI (target)        & $.970 {\scriptstyle\pm .006}$ & $.957 {\scriptstyle\pm .009}$ & $.469 {\scriptstyle\pm .080}$ & $.296 {\scriptstyle\pm .062}$ & $.892 {\scriptstyle\pm .004}$ & $.793 {\scriptstyle\pm .011}$ \\
GXD (target)        & $.996 {\scriptstyle\pm .000}$ & $.984 {\scriptstyle\pm .003}$ & $.980 {\scriptstyle\pm .002}$ & $.560 {\scriptstyle\pm .051}$ & $.926 {\scriptstyle\pm .003}$ & $\mathbf{.803} {\scriptstyle\pm .012}$ \\
GXD (all-logit)     & $\mathbf{.999} {\scriptstyle\pm .000}$ & $\mathbf{.984} {\scriptstyle\pm .003}$ & $\mathbf{.997} {\scriptstyle\pm .000}$ & $\mathbf{.568} {\scriptstyle\pm .050}$ & $\mathbf{.937} {\scriptstyle\pm .002}$ & $.783 {\scriptstyle\pm .015}$ \\
\bottomrule
\end{tabular}
\end{table}

For smooth activations and residual architectures, zero is no longer the inactive state and GXI's reference becomes misaligned with the actual reset intervention. In the SiLU MLP, GXI's Shock@5\% is comparable to Contribution and substantially worse than GXD (Figure~\ref{fig:app_shock5_variants}, center), and its Spearman $\rho$ on logit $L_1$ drops to $.469$ versus $.980$ for GXD (Table~\ref{tab:app_spearman_variants}). The same pattern appears in the ResNet-18, where GXI's Shock@5\% nearly matches Contribution while GXD approaches the oracle (Figure~\ref{fig:app_shock5_variants}, right). Figure~\ref{fig:app_pmnist_gxi_grad} shows this gap carries over to online continual learning: GXI matches GXD on ReLU but degrades on Tanh, SiLU, and Leaky ReLU, mirroring the failure mode of activation-based utilities that assume a zero baseline. This is an important ablation because it shows our addition of output sensitivity is not a meaningful improvement in our explored settings alone, and requires the intervention aware difference from reference component. 

\section{Extended Figures}
\label{app:extended_figures}

\begin{figure}[H]
  \centering
  \includegraphics[height=1.6in]{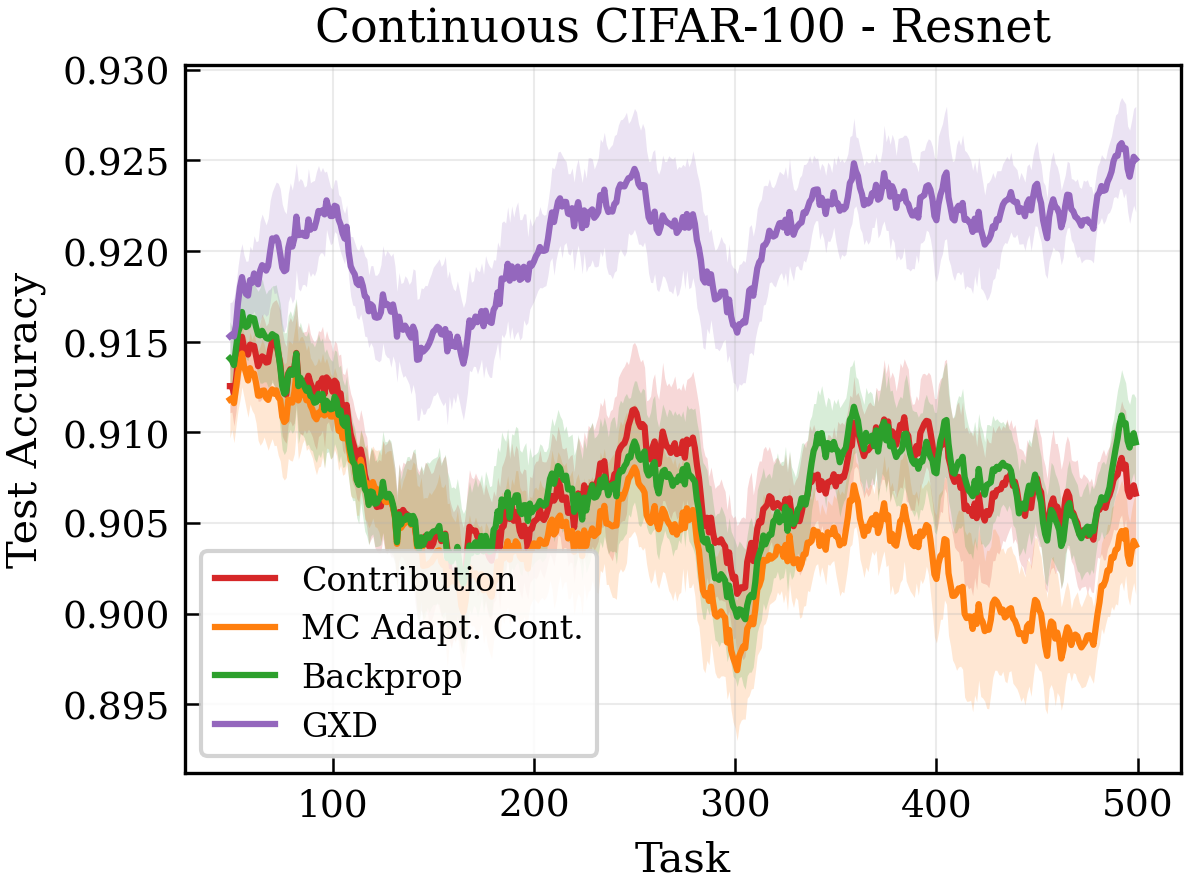}
  \caption{Absolute test accuracy on Continuous CIFAR-100 with ResNet.}
  \label{fig:app_cifar_abs}
\end{figure}

\end{document}